\documentclass{article}
\usepackage{ijcai25}

\usepackage{times}
\usepackage{soul}
\usepackage{url}
\usepackage[hidelinks]{hyperref}
\usepackage[utf8]{inputenc}
\usepackage[small]{caption}
\usepackage{graphicx}
\usepackage{amsmath}
\usepackage{amsthm}
\usepackage{booktabs}
\usepackage{algorithm}
\usepackage{algorithmic}
\usepackage[switch]{lineno}
\usepackage{forest} 
\usepackage{xcolor} 
\usepackage{adjustbox}
\usepackage{xcolor}

\title{A Survey on Data Curation for Visual Contrastive Learning: \\ Why Crafting Effective Positive and Negative Pairs Matters}

\author{
Shasvat Desai$^1$
\and
Debasmita Ghose$^2$\And
Deep Chakraborty$^{3}$
\affiliations
$^1$Independent Researcher\\
$^2$Yale University\\
$^3$University of Massachusetts Amherst\\
\emails
shasvat.desai@gmail.com, 
debasmita.ghose@yale.edu,
dchakraborty@umass.edu
}

\begin{document}

\maketitle

\begin{abstract}
Visual contrastive learning aims to learn representations by contrasting similar (positive) and dissimilar (negative) pairs of data samples. The design of these pairs significantly impacts representation quality, training efficiency, and computational cost. A well-curated set of pairs leads to stronger representations and faster convergence.
As contrastive pre-training sees wider adoption for solving downstream tasks, data curation becomes essential for optimizing its effectiveness. 
In this survey, we attempt to create a taxonomy of existing techniques for positive and negative pair curation in contrastive learning, and describe them in detail.
We also examine the trade-offs and open research questions in data curation for contrastive learning.
\end{abstract}

\section{Introduction}
\label{sec: Intro}

Contrastive learning has emerged in recent times as the dominant approach to self-supervised learning. 
The main idea of contrastive learning is to leverage the fact that similar data samples (positive pairs) should be positioned closer together in the embedding space, while dissimilar data samples (negative pairs) should be pushed further apart. 
The construction of the set of positive and negative pairs, referred to as \mbox{\emph{data curation}} in this paper, directly influences the informativeness of the embeddings \cite{khosla2020supervised,unremix,huynh2022boosting}. 

Data curation is a major challenge for contrastive learning as selecting ineffective data pairs may result in suboptimal embeddings, leading to poor generalization on downstream tasks. It also impedes the training process, resulting in higher training time and computational overhead, especially for large datasets. Effective data curation helps in training inference-efficient models \cite{udandarao2024active,evans2024data}, and accelerates training and model convergence \cite{xu2023cit,zhou2021cupid}.
Data curation can thus alleviate the issue of suboptimal representations and model convergence in the following ways: (1) Increasing diversity in data pairs, ensuring optimality of the learned embeddings that are invariant to intra-class variations, (2) Ensuring the selected data pairs are relevant and semantically aligned to prevent noisy samples from hindering training and enabling faster model convergence, (3) Enabling better alignment between the learned representations and the downstream task.

\begin{figure*}[h!]
    \centering
    \includegraphics[width=\linewidth]{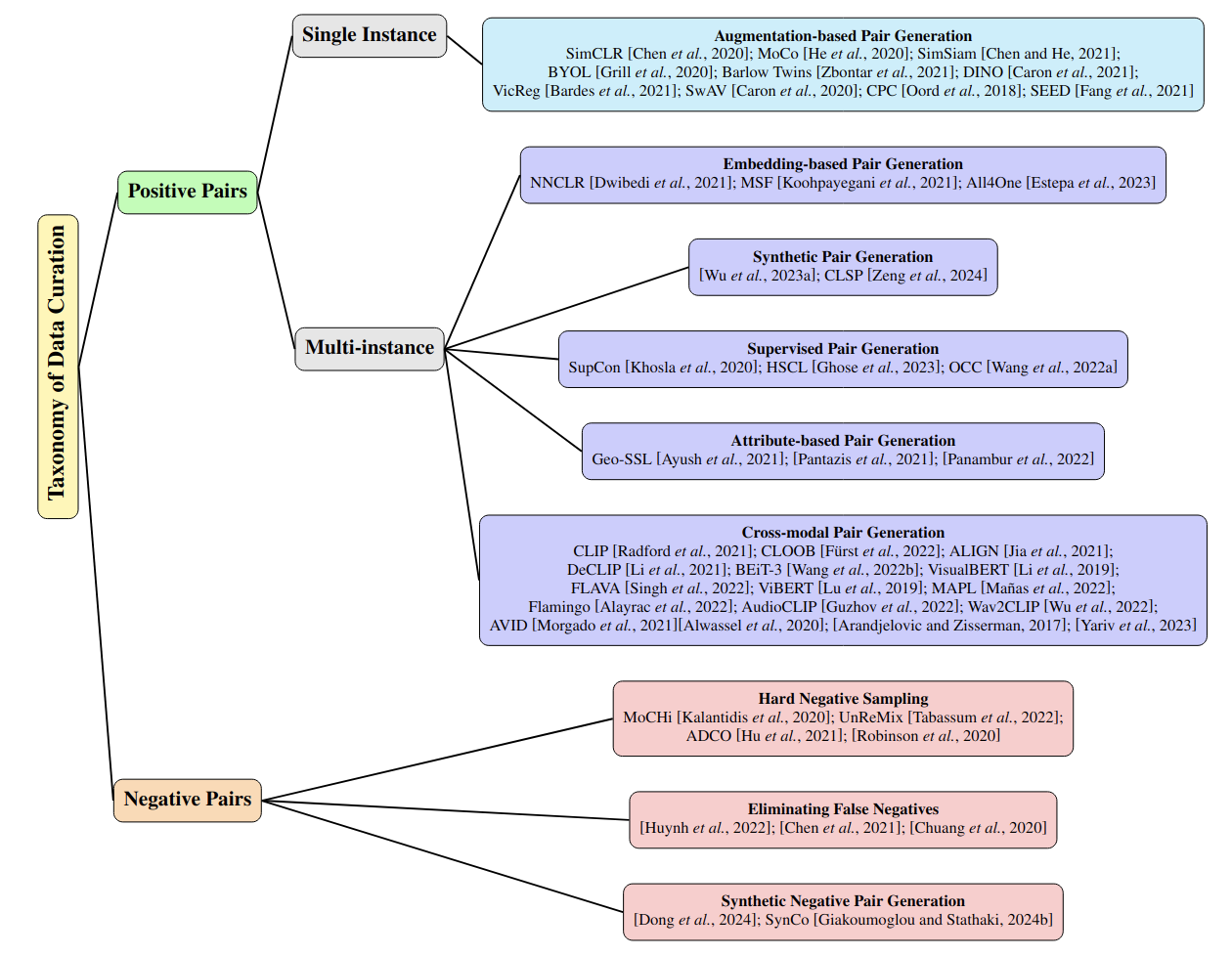}
          \caption{Taxonomy for crafting positive and negative pairs.}
\label{fig:taxonomy}
\end{figure*}

Data curation for contrastive learning can be approached in two ways: positive pair curation and negative pair curation. 
In the positive data curation direction, several recent works \cite{dwibedi2021little,wu2023synthetic,ayush2021geography} show that carefully designed positive pairs introduce diverse variations leading to generalized and informative embeddings. These techniques create positive pairs from different data samples (as opposed to different views of the same data sample) that are semantically aligned. This semantic alignment is defined through some similarity metric in embedding space \cite{koohpayegani2021mean,estepa2023all4one}, label, or available category information which can either be pre-defined \cite{khosla2020supervised}, or queried from an oracle or human in an online manner \cite{ghose2023tailoring}.  This semantic alignment reduces noise in the learning process and ensures relevance of pairs, while diversity ensures robustness, and together they optimize representation learning and improve model convergence. Similarly, some methods \cite{ayush2021geography} enable better alignment through leveraging predefined attributes based on the input domain or downstream task. Yet other works \cite{wu2023synthetic,zeng2024contrastive} propose using synthetically generated positives to generate diverse pairs that can potentially help with category imbalance, rare scenarios and modalities. 

Negative pairs in contrastive learning are often randomly sampled without explicit curation, but recent research highlights the importance of curating negatives to ensure semantic alignment and diversity through techniques like hard negative selection \cite{unremix}, synthetic negative generation \cite{giakoumoglou2024synco}, and false negative elimination \cite{huynh2022boosting}.
If negative pairs are too different from positives (easy negatives), they provide little learning signal—for example, distinguishing dogs from aircraft and ships is trivial. Instead, curating hard negatives that are closer to positives in embedding space \cite{robinson2020contrastive}, or incorporating adversarial negatives \cite{hu2021adco}, can improve representation learning. Additionally, synthetically generated negatives \cite{giakoumoglou2024synco,dong2024synthetic} enhance diversity, while methods like false negative elimination \cite{huynh2022boosting,chuang2020debiased} prevent semantically similar negatives from introducing conflicting learning signals. 

Thus, given the widespread use of contrastive pre-training, it is crucial to carefully select the data fed into the learning process to obtain robust and informative representations. 
Most existing surveys (\cite{gui2024survey,jaiswal2020survey,giakoumoglou2024review}) on visual contrastive learning focus on comparing architectural choices or learning objectives such as momentum encoders in MoCo \cite{he2020momentum} or stop-gradient mechanisms in SimSiam \cite{chen2021exploring}. In this survey, we take a complementary perspective by examining the role of data curation in contrastive learning. We discuss the tradeoffs between popular techniques, and finally pose some open questions for researchers interested in this direction.


\section{Taxonomy of Approaches for Positive and Negative Pair Curation}
\label{sec: Taxonomy}

In contrastive learning, a commonly used loss function is the InfoNCE loss defined below. It pulls similar (positive) pairs together while pushing dissimilar (negative) pairs apart in the embedding space and computes the similarity between an anchor and its positive counterpart, using a softmax over similarity scores. 

\begin{math}
\hspace{-1em}
\centering
 \mathcal{L} = 
- \frac{1}{N} \sum_{i=1}^{N} \log 
\frac{
\exp\left(\text{sim}\left(\mathbf{z}_i, \mathbf{z}_i^+\right)/\tau\right)
}{
\exp\left(\text{sim}\left(\mathbf{z}_i, \mathbf{z}_i^+\right)/\tau\right) 
+ \sum_{j=1}^{N} 
\exp\left(\text{sim}\left(\mathbf{z}_i, \mathbf{z}_j\right)/\tau\right)}
\end{math}

where \( \mathbf{z}_i \) is the representation of the anchor sample and \( \mathbf{z}_i^+ \) is the representation of the positive sample obtained through augmenting the same instance or using a criterion to select another instance. \( \mathbf{z}_j \) represents all samples in the batch (including negatives). \( \text{sim}(\cdot, \cdot) \) denotes the similarity function (commonly cosine similarity). \( \tau \) is the temperature scaling parameter and \( N \) is the number of samples in the batch.

\subsection{Positive Pair Creation Taxonomy}
The taxonomy of Positive Pair Creation can be categorized into two main groups: \textbf{single-instance positives} and \textbf{multi-instance positives}, as shown in Fig. 1.


Single-instance positive pair creation generates pairs by applying augmentations (e.g., cropping, color changes, geometric transformations) to a single sample \cite{chen2020simple}.  However, this approach limits diversity, as random augmentations fail to capture viewpoint changes, object deformations, or semantically similar instances within the same class. As a result, the model's generalization depends heavily on the augmentation strategy, which may not fully capture the intrinsic variations needed for learning robust embeddings.


To overcome the limitations of single-instance pairs, multi-instance positive pair curation creates pairs from different data samples rather than augmented views of the same sample, leading to greater diversity \cite{dwibedi2021little}. As shown in Fig. 1, multi-instance curation techniques include: (1) Embedding-based, which selects semantically similar instances in embedding space; (2) Synthetic, which generates positive pairs using generative models; (3) Supervised, which uses human or oracle-labeled data; (4) Attribute-based, which groups samples based on spatial, temporal, or other object-based attributes.  and (5) Cross-modal, which associates samples across different modalities.  By ensuring higher diversity and semantic alignment, multi-instance positive pairs improve representation learning and align embeddings more effectively with downstream tasks.

\subsection{Negative Pair Creation Taxonomy}

In typical contrastive learning approaches, negative pairs are created from samples not used to create the positive pair without considering their semantic content. However, recent work \cite{huynh2022boosting} suggests that uncurated negatives may lead to false negatives, where semantically similar samples are incorrectly treated as negatives. An effective negative sample selection strategy should balance easy and hard negatives while maintaining representativeness. Based on these principles, negative pair curation can be categorized into three main approaches (Fig. 1): (1) Hard Negative Selection, which prioritizes difficult negatives close to the anchor in embedding space; (2) False Negative Elimination, which removes or reclassifies semantically similar false negatives; and (3) Synthetic Negatives, where generative models create diverse, controlled negative samples. There is a subtle trade-off between (1) and (2). Hard negatives improve discrimination but risk overfitting, while false negative elimination reduces noise but may mistakenly remove challenging yet valid negatives, weakening the representations.

\section{Crafting Effective Positive Pairs
}
\label{sec: Positive_Pair}

\begin{figure*}[h!]
    \centering
    \includegraphics[width=0.95\linewidth]{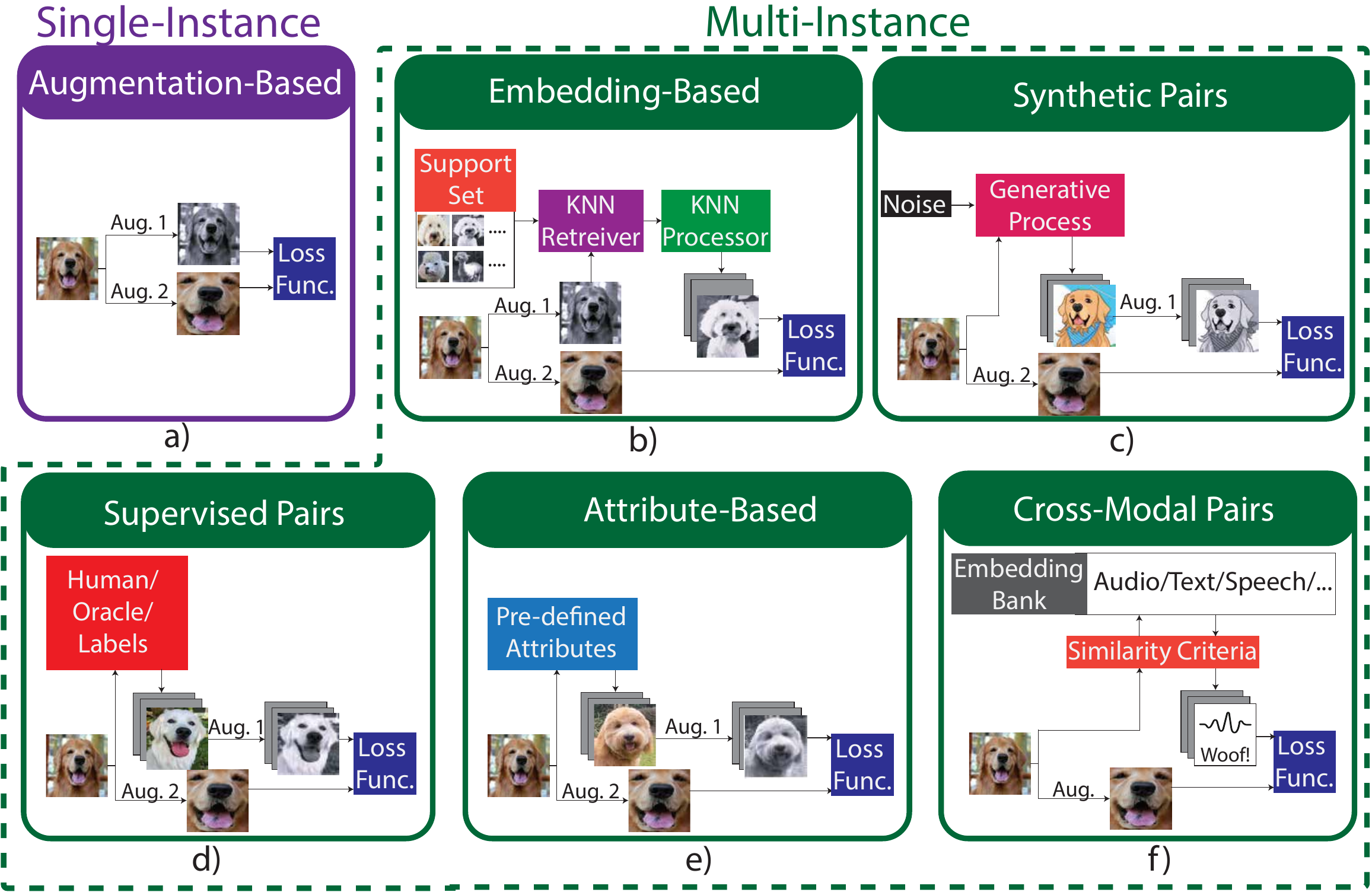}
    \caption{\textbf{Positive Pair Curation Techniques: } Positive pair selection can utilize single-instance and multi-instance techniques. (a) Single-instance curation applies augmentations to a single sample. On the other hand, multi-instance positive pair generation can be further classified into several category of techniques. (b) Embedding-based retrieves the top-K nearest neighbors of the anchor sample's augmentation in the embedding space and pairs them with other augmentations of the anchor. (c) Synthetic pairs generate data conditioned on the input, which is then augmented and paired with the augmented real sample (d) Supervised pairs use external sources (human labels, oracles, or annotations) to fetch another sample from the same category and create positive pairs. (e) Attributed-based: These methods group samples by shared attributes (e.g., golden retrievers paired with golden labrador retrievers based on fur color) and pair their respective augmentations. (f) Cross-modal: This involves creating semantically aligned pairs across multiple modalities. The figure shows image-text and speech-image pairing.}
    \label{fig:positive}
\end{figure*}

\subsection{Single Instance Positives}

This technique creates positive pairs using augmentations of a single sample without explicit curation, as shown in Fig. 2(a).  The negative pairs are also randomly sampled from the dataset and are uncurated. 

A common contrastive learning technique using this type of data curation is \textbf{SimCLR}\cite{chen2020simple}. SimCLR maximizes agreement between augmented views of the same data point using the InfoNCE loss, relying on large batch sizes to sample enough negatives. \textbf{MoCo} \cite{he2020momentum} addresses a drawback of SimCLR, which requires large negative samples that can be computationally expensive by using a momentum encoder and memory bank to maintain a queue of negatives dynamically. \textbf{SimSiam} \cite{chen2021exploring} eliminates the need for negative examples, using a stop-gradient mechanism to prevent representation collapse in its Siamese architecture. \textbf{BYOL} \cite{grill2020bootstrap} simplifies learning by aligning predictions from an online network with a momentum-maintained target network, achieving strong performance without negatives. \textbf{Barlow Twins} \cite{zbontar2021barlow} focuses on redundancy reduction by aligning embeddings and decorrelating feature dimensions, avoiding collapse naturally without negatives or momentum. \textbf{DINO} \cite{caron2021emerging} combines self-supervised learning with knowledge distillation using a teacher-student framework, producing generalized embeddings. \textbf{VicReg} \cite{bardes2021vicreg} introduces regularization to balance variance, invariance, and decorrelation in embeddings, ensuring quality without negatives or momentum encoders. \textbf{SwAV} \cite{caron2020unsupervised} uses clustering to align augmentations by mapping them to shared cluster assignments without direct contrastive loss. \textbf{CPC} \cite{oord2018representation} leverages contrastive loss in a latent space to predict future data segments, making it particularly effective for time-series tasks. Finally, \textbf{SEED} \cite{fang2021seed} simplifies training by using teacher-student distillation with pseudo-labels, reducing computational complexity. 

\subsection{Multi Instance Positives}
Multi-instance positive pair curation creates pairs from different samples rather than augmented views of the same sample.

\subsubsection{Embedding-Based Techniques}
Given an input candidate sample, this class of techniques first retrieves the K-nearest neighbors of one of the augmentations using a similarity metric in embedding space, as shown in Figure 2(b). Next, it uses the K-retrieved samples and the other augmentations of the candidate sample as positive pairs. 

Nearest-Neighbour Contrastive Learning of Visual Representations (NNCLR)\cite{dwibedi2021little} samples the nearest neighbors from the dataset in the latent space and treats them as positives. This provides
more semantic and intra-class variations to learn representations that are invariant to different viewpoints, deformations, and variations. The NNCLR framework relies entirely on a single nearest neighbor, limiting its potential. Mean Shift for Self-Supervised Learning (MSF) \cite{koohpayegani2021mean} addresses this limitation by proposing the use of \(k\) nearest neighbors to increase the diversity in the positive pairs. MSF shifts the embedding of each image to be closer to the \textit{mean} of the neighbors of its augmentation.
However, MSF is computationally expensive because the objective function must be computed \(k\) times for each neighbor. To address MSF's computational inefficiency, All4One \cite{estepa2023all4one} contrasts information from multiple neighbors by compiling information from the extracted \(k\) neighbors to create a pair of representations, called \textit{centroids}, which contain contextual information about all the neighbors. 

These techniques can be used when semantic clustering is needed for downstream applications. For instance, if the downstream task involves clustering similar faces, these techniques allow different views of the same person to be closer together, unlike single-instance positive techniques, which treat all other images as negatives.

\subsubsection{Synthetic Data Generation for Positive Pairs}

This class of techniques creates synthetic samples using a generative process conditioned on the candidate input sample. A positive pair is formed by combining the augmented generated sample with the augmentation of the original input sample, which is then processed by the encoder, as illustrated in Fig. 2(c).

Contrastive Learning with Synthetic Positives (CLSP) \cite{zeng2024contrastive} incorporates synthetic positives generated via a diffusion model. By interpolating Gaussian noise with diffusion-based features, CLSP creates images that resemble the anchor while varying the context and background, increasing diversity while preserving semantic meaning. Similarly, \cite{wu2023synthetic} introduces a GAN-based framework that dynamically generates hard positive pairs by jointly optimizing the GAN and contrastive model. However, this simultaneous training introduces instability and quality control challenges. These approaches are particularly useful in data-sparse scenarios, rare modalities, or domains where obtaining real data is challenging, such as cross-modal medical applications (e.g., speech-image pairs).

\subsubsection{Supervised Pairing Techniques}

These techniques use external data sources, such as human preferences, privileged information from an oracle, or an annotated dataset, to derive meaningful metadata and semantic categories and create positive pairs, as shown in Fig. 2(d).

Supervised contrastive learning (SupCon) \cite{khosla2020supervised} leverages ground truth labels to enhance representation learning by incorporating category-level supervision. Instead of defining positive pairs through augmentations of a single instance, SupCon creates positive pairs from multiple samples of the same category as the anchor, ensuring that representations capture category-level semantic similarities rather than just instance-specific features.

Building upon this, \cite{ghose2023tailoring} propose a method to create positive pairs on the fly by passively observing humans provide limited positive examples while working collaboratively with a robot without explicitly marking negatives. This aligns with Positive-Unlabeled (PU) Learning \cite{bekker2020learning}, where only positives are known, and the model infers meaningful distinctions. Contrastive learning then clusters these examples, ensuring representations align with human expectations in a task-adaptive manner.
Similarly, Oracle-guided Contrastive Clustering (OCC) \cite{wang2022oracle} uses a deep clustering framework designed to create positive pairs for contrastive loss by incorporating oracle feedback into the clustering process, ensuring that learned representations align with user-specific clustering preferences. Instead of relying purely on instance similarity in the embedding space, OCC actively queries an oracle (human or predefined rule) to determine whether two samples should belong to the same cluster.

These techniques are useful in scenarios when labeled data is available, and the goal is to cluster semantically similar items within the same class or the downstream task requires discrimination between subtle intra-class variations. Leveraging labels and semantic information to create pairs enables us to generate embeddings that are better aligned to the downstream application. Intuitively, these techniques should eliminate false negatives (through semantic clustering) and potentially reduce noise in the generated embeddings. 

\subsubsection{Attribute-based Pairing Techniques}

Attribute-based pairing entails selecting positive pairs based on task-specific criteria, as shown in Fig. 2(e). Attributes can be generic, such as \textit{``a golden colored object"} or more specific such as \textit{``dog with golden colored coat"}

For instance, Geography-aware self-supervised learning \cite{ayush2021geography} leverages spatial and temporal attributes to create temporal positive pairs from images of the same geographical location taken at different times. They demonstrate their approach in the Remote sensing domain because it is easy to obtain multiple geo-located images of the exact location over time. Similarly, \cite{pantazis2021focus} leverages the natural variations in sequential images from static cameras, utilizing contextual information such as spatial and temporal relationships to identify high-probability positive pairs—images likely depicting the same visual concept. Yet another approach proposed by \cite{panambur2022self} leveraged domain-specific attributes like geological structures, terrain textures, and spatial and scientific properties to form positive pairs for terrain categorization in Martian terrain.

These techniques are useful when domain-specific contextual attributes are known. However, uneven attribute distribution can lead to over-representation of certain pair types and hinder performance on unseen variations, so careful attribute selection and balancing are essential.

\subsubsection{Cross-modal Positive Pairing Techniques}
Cross-modal contrastive learning aims to learn meaningful representations across different data modalities  to improve performance in various tasks that involve multi-modal data,
as shown in Fig. 2(f). 

\noindent
\textbf{Image-Text Pairing: }
Image-text pairing aims to align visual and textual information to learn effective representations. 
CLIP \cite{radford2021learning}, CLOOB \cite{furst2022cloob}, ALIGN \cite{jia2021scaling} employs contrastive learning to learn shared representations by aligning visual and textual data which facilitate downstream tasks like zero-shot image classification and cross-modal retrieval. BEiT-3 \cite{wang2022image}, VisualBERT \cite{li1908visualbert}, FLAVA \cite{singh2022flava}, LXMERT \cite{tan2019lxmert} are a class of methods that introduce a unified masked data modeling objective. Given a partially masked caption, the objective is to predict the masked words based on the corresponding image. They learn representations that capture the relationships between images and texts by masking parts of the input and training the model to predict the missing information. MAPL \cite{manas2022mapl} and Flamingo \cite{alayrac2022flamingo} keep the pre-trained vision encoder and language model frozen to learn a lightweight mapping between their representation spaces, enabling few-shot learning with minimal parameter updates.

\noindent
\textbf{Audio-Image-Text Pairing: }
AudioCLIP \cite{guzhov2022audioclip}, Wav2CLIP \cite{wu2022wav2clip} learns audio representations by distilling knowledge from the CLIP model to jointly learn a shared representation of audio data alongside image and text modalities. CLAP \cite{wu2023large} trains a dual-encoder model to align audio and text embeddings. 

\begin{figure*}[h!]
    \centering
    \includegraphics[width=0.97\linewidth]{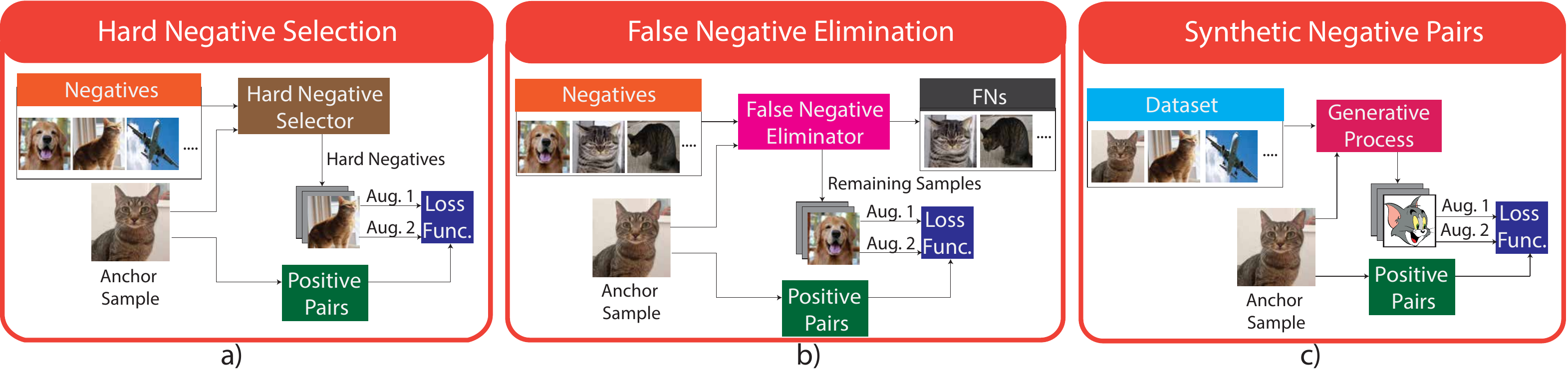}
    \caption{\textbf{Negative Pair Curation Techniques}: 
    This figure shows three categories of techniques for negative pair curation. (a). Hard Negative Selection prioritizes negatives that are semantically similar to the anchor sample, such as a different cat breed, instead of an unrelated category like an airplane. The negatives are then augmented and fed into the encoder. (b). False Negative Elimination removes or reclassifies negatives that are highly similar to the anchor sample, preventing the model from mistakenly separating highly similar samples. The remaining negatives are then augmented before encoding. Hard negatives improve discrimination but risk overfitting, while false negative elimination reduces noise but may mistakenly remove challenging yet valid negatives, weakening the representations. (c). Synthetic negative pairs are created by feeding the positive and negative samples(dataset) into a generative process and conditioned on the anchor sample to create realistic but distinct negatives. The generated samples then undergo augmentation and are fed with the positive pairs to the downstream encoder.}
    \label{fig:negative}
\end{figure*}



\noindent
\textbf{Audio-Visual Pairing: }
Audio-Visual Instance Discrimination (AVID) \cite{morgado2021audio} and \cite{alwassel2020self} emphasize cross-modal discrimination, aiming to align audio and visual features effectively. They use cross-modal clustering, where shared semantic clusters are learned across modalities by mapping audio and video representations into a joint embedding space. Building upon this, \cite{yariv2023audiotoken} adapts pre-trained text-conditioned diffusion models, like Stable Diffusion, by converting audio inputs into text-like embeddings through a learnable adapter. These embeddings serve as prompts for the diffusion model to generate audio-aligned images.

\section{Crafting Effective Negative Pairs}
\label{sec: Negative_Pair}

\subsection{Hard Negative Selection}

Hard negatives are those negative samples that are particularly similar to the anchor (the positive sample) in the embedding space, making them more likely to be misclassified. By incorporating such samples, the model is forced to refine its representation, learning more discriminative features to distinguish between fine-grained differences. These hard negatives are then fed into the encoder along with the positive pair, as shown in Fig. 3(a).

MoCHi, \textit{(M)ixing (o)f (C)ontrastive (H)ard negat(i)ves} \cite{Hardnegativemixing}, creates hard negatives by combining features of existing hard negatives in the embedding space. It identifies existing negatives for a given anchor that are most similar to the candidate positive sample in the embedding space and combines these hard negatives at the feature level to create synthetic negatives that are even closer to the anchor, increasing the difficulty of the contrastive task.

Uncertainty and Representativeness Mixing (UnReMix) \cite{unremix} selects negative samples based on three key properties. Anchor similarity ensures that negative samples closely resemble the anchor but belong to different classes. Model uncertainty prioritizes negative samples with higher prediction uncertainty, focusing the learning process on less confident regions of the data space. Representativeness emphasizes selecting negatives that reflect the overall data distribution rather than outliers. Similarly, \cite{robinson2020contrastive} samples negatives close to the anchor in the embedding space.  These negatives are generated adversarially or synthesized through feature interpolation, promoting fine-grained learning. A balanced mix of hard and easy negatives ensures stability and prevents overfitting during training.

Yet another approach \cite{hu2021adco} follows a min-max optimization framework, where the encoder minimizes the contrastive loss by learning to separate positives from negatives while the negative adversaries maximize the loss by generating challenging and indistinguishable negatives.

\subsection{Removal of False Negatives}
False negatives are samples from different images with the same semantic content, therefore they should hold certain similarity. Contrasting false negatives induces two critical issues in representation learning: 1) discarding semantic information and 2) slow convergence due to the addition of noise in the learning process. For instance, a cat's head in one image might be attracted to its fur (positive pair) but repelled from the similar fur in another image of a cat (negative pair), creating conflicting objectives. Eliminating false negatives involves taking a batch of negative samples and removing those highly similar to positives, as shown in Figure 3(b). The rest of the samples in the batch undergo augmentations and are sent to the encoder along with the positive pairs.

\cite{huynh2022boosting} introduces methods to identify these false negatives and proposes two strategies to mitigate their impact: elimination and attraction. Elimination identifies and excludes potential false negatives from the negative sample set, preventing the model from learning misleading distinctions. In contrast, false negative attraction reclassifies them as positives(makes them true positives), encouraging the model to learn representations that acknowledge their semantic similarity. Similarly, \cite{chen2021incremental} dynamically detects false negatives based on semantic similarity and reclassifies them as positives, thus reducing noise in the learning process. 

\cite{chuang2020debiased} takes a different approach to mitigate the impact of false negatives in contrastive learning by introducing a re-weighted loss function. This loss adjusts the contribution of each negative sample based on its likelihood of being a true negative without requiring label information. The approach improves representation learning by minimizing the influence of false negatives, achieving better performance in self-supervised settings across various domains.
These techniques help ensure the negative pairs are relevant and the generated embeddings are aligned to the downstream task.

\subsection{Synthetic Hard Negatives}
Synthetic negatives can be created using various techniques, including generative models, feature space interpolation, or rule-based algorithms that modify existing data. Once created, their augmentation and positive pairs are sent to the encoder, as shown in Fig. 3(c).

\textit{Synthetic Hard Negative Samples for Contrastive Learning} \cite{dong2024synthetic} involves mixing existing negative samples in the feature space to create more challenging negatives synthetically. To address the issue of false negatives—samples incorrectly labeled as negative but semantically similar to the anchor, this work incorporates a debiasing mechanism, ensuring the model focuses on truly dissimilar negative samples. 
The selected hard negatives are then combined through linear interpolation to create synthetic negative samples that are even closer to the anchor in the feature space. 

Similarly, another approach \cite{giakoumoglou2024synco} builds upon the MoCo framework \cite{he2020momentum} to create diverse synthetic hard negatives on the fly with minimal computational overhead. It generates negatives by interpolating between positive and negative samples in the feature space, extrapolating beyond the positive sample in the direction of a negative sample, applying small perturbations to positive samples to generate negatives, and using adversarial methods to craft indistinguishable negatives.
\section{Discussion}



\subsection{Trade-offs Between Techniques that Generate Positive Pairs from Multiple Instances}

Embedding-based positive pairs leverage semantic similarity in the embedding space to capture fine-grained variations, enabling models to learn intricate features such as fur color or ear shape when distinguishing dog breeds. However, this approach can be computationally intensive and prone to noisy embeddings, particularly in the early training stages.

Synthetic data generation dynamically creates challenging and diverse positive pairs. However, these models 
require careful tuning to prevent degradation from low-quality synthetic samples. A significant challenge is the visual fidelity of synthetic samples, as they might lack the richness and detail of real-world data, resulting in positive pairs that do not fully capture the desired semantic similarity required for a given downstream task. 
Another concern is semantic misalignment, where synthetic samples may inadvertently introduce artifacts or distortions that diverge from real-world semantics that can lead to representations that overfit to synthetic peculiarities. Addressing domain gap challenges requires novel strategies like combining synthetic positives with real data, using hybrid training approaches, or incorporating domain adaptation methods that can help bridge the gap.

Supervised pairing leverages label information to create positive pairs. This technique is particularly beneficial when labeled data is available, and class-specific clustering is essential for a downstream task. It is helpful for tasks such as class-specific retrieval, where the goal is to fetch semantically similar items within the same class, or tasks requiring discrimination between subtle intra-class variations. The biggest drawback is that this method assumes the availability of labels, which might not be feasible in every scenario.

Attribute-based pairing leverages contextual attributes and is effective in scenarios where domain-specific context plays a crucial role in learning robust representations. One challenge with this method is that attributes may be unevenly distributed across the dataset, leading to over-representing certain pair types. Another challenge is that models may overfit to specific attribute values instead of learning generalizable representations. 

Cross-modal pair generation is particularly useful for multimodal learning, which is becoming increasingly prevalent. However, this method depends on the availability of semantically aligned multimodal pairs. Misaligned pairs can result in less meaningful representations. Another significant challenge is obtaining aligned data for modalities that can be paired for contrastive learning, mainly when one of the modalities is rare or difficult to generate labels for.

\subsection{Trade-offs Between Techniques that Generate Negative Pairs from Multiple Instances}

Hard negative selection is valuable for generating informative embeddings. However, overemphasizing hard negatives can lead to overfitting, where the model learns to differentiate subtle, unimportant variations rather than capturing meaningful representations.  While hard negatives provide valuable gradients, easy negatives ensure stability and prevent overfitting to challenging examples. Hence, creating the right mix of hard and easy negatives in a batch for learning is essential.

Eliminating false negatives helps reduce noise, but accurately defining them is challenging. Over-aggressive removal or reclassification as positives can reduce diversity and make negatives too easy, weakening contrastive learning. Conversely, being too conservative allows false negatives to persist, hindering learning. The key is to balance diversity and difficulty, ensuring negatives remain challenging (hardness) without incorrectly reclassifying true negatives as positives. 

Synthetic negatives offer a scalable approach for generating diverse negative pairs, but they face challenges similar to synthetic positive generation, including semantic misalignment and domain gaps. Additionally, maintaining a large pool of negatives or dynamically synthesizing new ones incurs significant computational costs, making efficiency a key consideration in their implementation. 

\section{Open Questions}

\subsection{Balancing Diversity and Relevance in Pairs}
Diversity can be defined as the variation in data samples and pairs used in training while relevance is the level of semantic alignment between the pairs. Ideally, there should be high semantic alignment across pairs of the same category and low alignment across positive and negative categories.

Balancing diversity and relevance is a critical challenge. Both factors are essential: diversity ensures robustness and generalization, while semantic alignment guarantees meaningful and task-relevant representations. Diverse positive pairs may include instances that, while related, are semantically weakly aligned and hence irrelevant, leading to noisy training process and embeddings. Conversely, tightly aligned pairs might be more semantically aligned and relevant but miss significant variations and lack diversity.
Diversity is preferred when the downstream task is unknown, while relevance is prioritized when strong task alignment is needed. 
Future research can focus on dynamic training approaches---
Should early phases focus on introducing diversity to learn general features, followed by more task-aligned features? Or should specialized features be learned first from task-relevant data, and learning generalizable features relegated to later phases?

\subsection{Dealing with Emerging Modalities in Contrastive Learning}

As new modalities—such as LiDAR, hyperspectral imaging, and haptic feedback, become prominent in various applications, the challenge of integrating these modalities into contrastive learning frameworks emerges \cite{dai2024advancing}. 
Many emerging modalities suffer from the lack of large-scale labeled or even unlabeled datasets. They often exhibit high noise levels or variability due to environmental factors or inherent measurement inaccuracies. Moreover, unlike text or images, pre-trained models for emerging modalities are rare, making initialization more challenging. Future research can focus on scalable and efficient strategies to handle the diversity and complexity of these new data types.


\newpage
\bibliographystyle{named}
\bibliography{ijcai25}

\end{document}